\documentclass[11pt,a4paper]{article}
\usepackage[hyperref]{emnlp-ijcnlp-2019}
\usepackage{times}
\usepackage{latexsym}

\usepackage{url}
\usepackage{graphicx}
\usepackage{wrapfig}
\usepackage{amsmath}
\usepackage{amssymb}
\usepackage{color,xcolor,colortbl}
\usepackage{enumitem}
\usepackage{algorithm}
\usepackage{algorithmic}
\usepackage{bm,bbm}
\usepackage{booktabs}
\usepackage{mathtools}
\usepackage{array}
\usepackage{subcaption}
\usepackage{pbox}
\usepackage{pifont}
\usepackage{relsize}

\newcommand\numberthis{\addtocounter{equation}{1}\tag{\theequation}}

\definecolor{darkblue}{rgb}{0.0, 0.0, 0.55}
\newenvironment{fontppl}{\fontfamily{ppl}\selectfont}{\par} 
\definecolor{input}{rgb}{0.63, 0.79, 0.95}
\definecolor{conv1}{rgb}{1.0, 0.89, 0.77}
\definecolor{maxpool}{rgb}{0.92, 0.3, 0.26}
\definecolor{dense}{rgb}{0.82, 0.62, 0.91}
\definecolor{relu}{rgb}{0.76, 0.33, 0.76}

\usepackage{multirow}
\usepackage{comment}

\aclfinalcopy 

\title{Multi-Document Summarization with Determinantal Point Processes and Contextualized Representations}

\author{Sangwoo Cho$^\spadesuit$, Chen Li$^\diamondsuit$, Dong Yu$^\diamondsuit$, Hassan Foroosh$^\spadesuit$, Fei Liu$^\spadesuit$\\[0.8em]
$^\spadesuit$Computer Science Department, University of Central Florida\\
$^\diamondsuit$Tencent AI Lab, Bellevue, WA, USA\\[0.6em]
{\tt\relscale{0.97} \textsf{swcho@knights.ucf.edu} 
\textsf{\{ailabchenli, dyu\}@tencent.com}
\textsf{\{foroosh,feiliu\}@cs.ucf.edu}}\\
}

\date{}

\begin{document}
\maketitle
\begin{abstract}

Emerged as one of the best performing techniques for extractive summarization, determinantal point processes select the most probable set of sentences to form a summary according to a probability measure defined by modeling sentence prominence and pairwise repulsion.
Traditionally, these aspects are modelled using shallow and linguistically informed features, but the rise of deep contextualized representations raises an interesting question of whether, and to what extent, contextualized representations can be used to improve DPP modeling.
Our findings suggest that, despite the success of deep representations, it remains necessary to combine them with surface indicators for effective identification of summary sentences.

\end{abstract}

\section{Introduction}
\label{sec:intro}

Determinantal point processes, shortened as DPP, is one of a number of optimization techniques that perform remarkably well in summarization competitions~\cite{Hong:2014}.
These optimization-based summarization methods include integer linear programming~\cite{Gillick:2009:NAACL}, minimum dominating set~\cite{Shen:2010}, maximizing submodular functions under a budget constraint~\cite{Lin:2010:NAACL,Yogatama:2015:EMNLP}, and DPP~\cite{Kulesza:2012}.
DPP is appealing to extractive summarization, since not only has it demonstrated promising performance on summarizing text/video content~\cite{Gong:2014,Zhang:2016:DPP,Sharghi:2018}, but it has the potential of being combined with deep neural networks for better representation and selection~\cite{Gartrell:2018}.

The most distinctive characteristic of DPP is its decomposition into the \emph{quality} and \emph{diversity} measures~\cite{Kulesza:2012}.
A \emph{quality} measure is a positive number indicating how important a sentence is to the extractive summary.
A \emph{diversity} measure compares a pair of sentences for redundancy.
If a sentence is of high quality, any \emph{set} containing it will have a high probability score.
If two sentences contain redundant information, they cannot both be included in the summary, thus any \emph{set} containing both of them will have a low probability.
DPP focuses on selecting the most probable \emph{set} of sentences to form a summary according to sentence quality and diversity measures.

To better measure quality and diversity aspects, we draw on deep contextualized representations. 
A number of models have been proposed recently, including ELMo~\cite{Peters:2018}, BERT~\cite{Devlin:2018}, XLNet~\cite{Yang:2019,Dai:2019}, RoBERTa~\cite{Liu:2019:RoBERTa} and many others.
These representations encode a given text into a vector based on left and right context.
With carefully designed objectives and billions of words used for pretraining, they have achieved astonishing results in several tasks including predicting entailment relationship, semantic textual similarity, and question answering.
We are particularly interested in leveraging BERT for better sentence quality and diversity estimates.

This paper extends on previous work~\cite{Cho:2019} by incorporating deep contextualized representations into DPP, with an emphasis on better sentence selection for extractive multi-document summarization.
The major research contributions of this work include the following: 
(\emph{i}) we make a first attempt to combine DPP with BERT representations to measure sentence quality and diversity and report encouraging results on benchmark summarization datasets;
(\emph{ii}) our findings suggest that it is best to model sentence \emph{quality}, i.e., how important a sentence is to the summary, by combining semantic representations and surface indicators of the sentence, whereas pairwise sentence \emph{dissimilarity} can be determined by semantic representations only;
(\emph{iii}) our analysis reveals that combining contextualized representations with surface features (e.g., sentence length, position, centrality, etc) remains necessary, as deep representations, albeit powerful, may not capture domain-specific semantics/knowledge such as word frequency.

\section{DPP for Summarization}
\label{sec:dpp}

Determinantal point process (Kulesza and Taskar, 2012\nocite{Kulesza:2012}) defines a probability measure $\mathcal{P}$ over all subsets ($2^{|\mathcal{Y}|}$) of a ground set containing all document sentences $\mathcal{Y} = \{1,2,\cdots,\textsf{N}\}$. 
Our goal is to identify a most probable subset $Y$, corresponding to an extractive summary, that achieves the highest probability score. 
The probability measure $\mathcal{P}$ is defined as
\begin{align*}
\mathcal{P}(Y;L) &= \frac{\mbox{det}(L_Y)}{\mbox{det}(L + I)}, 
\numberthis\label{eq:p_y}\\
\sum_{Y \subseteq \mathcal{Y}}\mbox{det}(L_Y) &= \mbox{det}(L + I),
\numberthis\label{eq:sum_det_y}
\end{align*}
where 
$\mbox{det}(\cdot)$ is the determinant of a matrix;
$I$ is the identity matrix;
$L \in \mathbb{R}^{\textsf{N} \times \textsf{N}}$ is a positive semi-definite (PSD) matrix, known as the $L$-ensemble; 
$L_{ij}$ indicates the correlation between sentences $i$ and $j$;
and $L_Y$ is a submatrix of $L$ containing only entries indexed by elements of $Y$.
As illustrated in Eq.~(\ref{eq:p_y}), the probability of an extractive summary $Y \subseteq \mathcal{Y}$ is thus proportional to the determinant of the matrix $L_Y$.

Kulesza and Taskar~\shortcite{Kulesza:2012} introduce a decomposition of the $L$-ensemble matrix: 
$L_{ij} = q_i \cdot S_{ij} \cdot q_j$
where 
$q_i \in \mathbb{R}^+$ is a positive number indicating the \emph{quality} of a sentence
and $S_{ij}$ is a measure of \emph{similarity} between sentences $i$ and $j$.
The $q$ and $S$ model the sentence quality and pairwise similarity respectively and contribute to the $L$-ensemble matrix.
A log-linear model is used to determine sentence quality: $q_i = \mbox{exp}(\boldsymbol{\theta}^\top \mathbf{f}(i))$, where $\mathbf{f}(i)$ is a feature vector for sentence $i$ and $\boldsymbol{\theta}$ are feature weights to be learned during DPP training. 
We optimize $\boldsymbol{\theta}$ by maximizing log-likelihood with gradient descent, illustrated as follows:
{\medmuskip=1mu
\thinmuskip=1mu
\thickmuskip=1mu
\nulldelimiterspace=0pt
\scriptspace=0pt
\begin{align*}
\mathcal{L}(\boldsymbol\theta) = \sum_{m=1}^M \log \mathcal{P}(\hat{Y}^{(m)}; L^{(m)}(\boldsymbol{\theta})),
\numberthis\\
\nabla_{\boldsymbol\theta} = \sum_{m=1}^M \sum_{i \in \hat{Y}^{(m)}}\mathbf{f}(i) - \sum_j \mathbf{f}(j)K_{jj}^{(m)},
\numberthis\label{eq:mle}
\end{align*}}

\noindent where $M$ is the total number of training instances; 
$\hat{Y}^{(m)}$ is the ground-truth summary of the $m$-th instance; 
$K=L(L+I)^{-1}$ is the kernel matrix and $\mathcal{P}(\hat{Y}^{(m)}; L^{(m)}(\boldsymbol{\theta}))$ is defined by Eq.~(\ref{eq:p_y}).
We refer the reader to~\cite{Kulesza:2012} for details on gradient derivation (Eq.~(\ref{eq:mle})). 
In the following we describe two BERT models to respectively estimate sentence pairwise similarity and importance.
The trained models are then plugged into the DPP framework for computing $S$ and $q$.

\begin{figure}[!tb]
\centering
\includegraphics[width=1.0\linewidth]{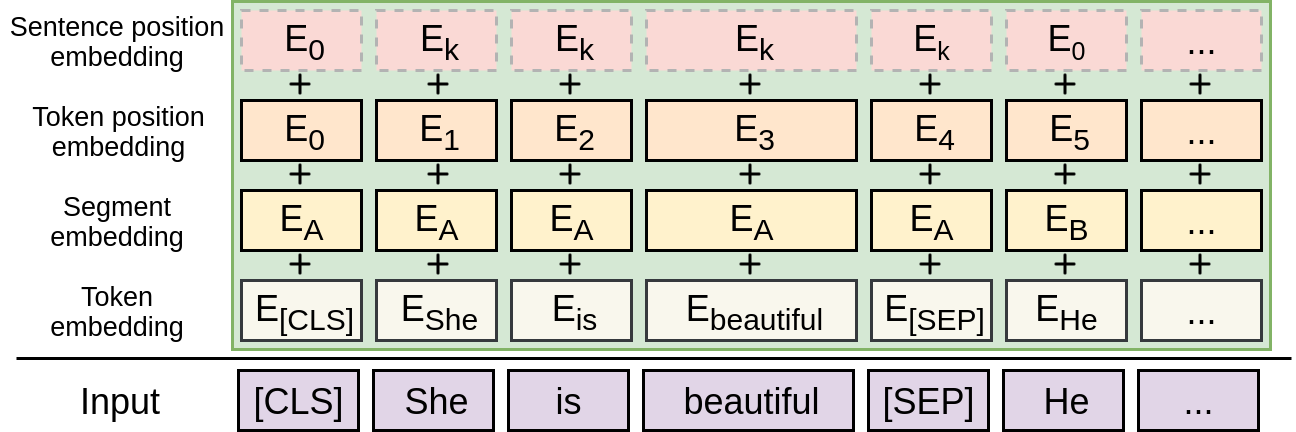}
\caption{
BERT-\emph{sim} and BERT-\emph{imp} utilize embeddings for tokens, segments, token position in a sentence and sentence position in a document.
These embeddings are element-wisely added up then fed into the model.
}
\label{fig:bert_embedding}
\end{figure} 

\begin{table}[t]
\setlength{\tabcolsep}{5pt}
\renewcommand{\arraystretch}{1.2}
\centering
\begin{small}
\begin{tabular}{|l|rrr|}
\hline
CNN/DM & mean & min & max \\
\hline
train-pos  & 13.95 & 1 & 318\\
train-neg  & 21.90 & 1 & 337\\
\hline
\hline
DUC-04 & 2.22 & 1 & 5 \\
TAC-11 & 1.67 & 1 & 5 \\
\hline
\end{tabular}
\end{small}
\caption{Position of summary-worthy sentences in a document for single-doc (CNN/DM) and multi-doc datasets (DUC-04, TAC11). 
`pos' are summary-worthy document sentences; 
`neg' are sentences that are randomly sampled from the same document.
}
\label{tab:sum_sent_pos}
\end{table}

\subsection{BERT Architecture}
\label{sec:bert_models}

We introduce two models that fine-tune the BERT-base architecture~\cite{Devlin:2018} to calculate the similarity between a pair of sentences (BERT-\emph{sim}) and learn representations that characterize the importance of a single sentence (BERT-\emph{imp}). 
Importantly, training instances for both BERT models are derived from \emph{single-document} summarization dataset~\cite{Hermann:2015} by Lebanoff et al.~\shortcite{Lebanoff:2019}, containing a collection of single sentences (or sentence pairs) and their associated labels.
During testing, the trained BERT models are applied to single sentences and sentence pairs derived from \emph{multi-document} input to obtain quality and similarity measures.

BERT-\emph{sim} takes as input a pair of sentences and transforms each token in the sentence into an embedding using an embedding layer.
They are then passed through the BERT-base architecture to produce a vector representing the input sentence pair. 
The vector, denoted by $\mathbf{u} \in \mathbb{R}^d$, is the final hidden state corresponding to the ``{\footnotesize\textsf{[CLS]}}'' token ($d$=768), which is used as the aggregate sequence representation.
$\mathbf{u}$ is passed through a feed-forward layer with the same dimension $d$, followed by a dropout layer, and a final softmax prediction layer to classify whether a pair of sentences contain redundant information or not. 
Once the model is trained, we can apply it to a pair of sentences $i$ and $j$ to obtain the similarity score $S_{ij}$.

BERT-\emph{imp} uses a similar architecture to predict if any single sentence is important to the summary. 
Once the model is trained, we can apply it to the $i$-th sentence to generate a vector $\mathbf{u}_i$ which is used as the feature representation $\mathbf{f}(i)$ for the $i$-th sentence when computing $q_i$.

The embedding layer, illustrated in Fig.~\ref{fig:bert_embedding}, consists of several types of embeddings, respectively representing tokens, segments, the token position in a sentence and sentence position within a given document. 
These embeddings are element-wisely added up then fed to the model.
The sentence position embeddings are incorporated in this work to capture the position of a sentence in the article.
It is utilized only by BERT-\emph{imp}, as position matters for sentence importance but not quite so for pairwise similarity. 
As shown in Table~\ref{tab:sum_sent_pos}, positive sentences in the training data (see \S\ref{sec:data}) tend to appear at the beginning of an article, consistently more so than negative sentences. 
Further, ground-truth summary sentences of the DUC and TAC datasets are likely to appear among the first five sentences of an article, indicating position embeddings are crucial for training the BERT-\emph{imp} model.

\subsection{DPP Training}
DPP training focuses on estimating the weights of features used in $q_i = \mbox{exp}(\boldsymbol{\theta}^\top \mathbf{f}(i))$, which is a log-linear model used for computing sentence quality.
The sentence similarity scores $S_{ij}$ are produced by BERT-\emph{sim}; they do not change during DPP training.
We obtain contextualized representations for the $i$-th sentence, i.e., $\mathbf{f}(i) \in \mathbb{R}^d$, from the penultimate layer ($\mathbf{u}_i$) of BERT-\emph{imp}.

In addition, a number of surface indicators\footnote{The sentence features include the length and position of a sentence, the cosine similarity between sentence and document TF-IDF vectors~\cite{Kulesza:2011}. We abstain from using sophisticated features to avoid model overfitting.}, denoted by $\mathbf{v}_i \in \mathbb{R}^{d'}$, are extracted for sentence $i$. 
To combine surface indicators and contextualized representations, we concatenate $\mathbf{u}_i$ and $\mathbf{v}_i$ as sentence features.
We also take a weighted average\footnote{The coefficient is set to be 0.9 for both datasets.} of $S_{ij}$ and $C_{ij}$ as an estimate of pairwise sentence similarity, where $C_{ij}$ is the cosine similarity of sentence TF-IDF vectors.
DPP training learns feature weights $\boldsymbol{\theta} \in \mathbb{R}^D$, where $D = d+d'$ if the sentence features are concatenated, otherwise $D = d$.
DPP is trained on multi-document summarization data with gradient descent (Eq.~(\ref{eq:mle})).

\begin{table}[!t]
\setlength{\tabcolsep}{5pt}
\renewcommand{\arraystretch}{1.2}
\centering
\begin{small}
\begin{tabular}{|l|rrr|}
\hline
& \multicolumn{3}{c|}{\textbf{DUC-04}}\\
\textbf{System} & \textbf{R-1} & \textbf{R-2} & \textbf{R-SU4} \\
\hline
\hline
Opinosis{\scriptsize~\cite{Ganesan:2010}} & 27.07 & 5.03 & 8.63\\
Extract+Rewrite{\scriptsize~\cite{Song:2018}} & 28.90 & 5.33 & 8.76 \\
Pointer-Gen{\scriptsize~\cite{See:2017}} & 31.43 & 6.03 & 10.01\\
SumBasic{\scriptsize~\cite{Vanderwende:2007}} & 29.48 & 4.25 & 8.64\\
KLSumm{\scriptsize(Haghighi et al., 2009)\nocite{Haghighi:2009}} & 31.04 & 6.03 & 10.23 \\
LexRank{\scriptsize~\cite{Erkan:2004}} & 34.44 & 7.11 & 11.19 \\
ICSISumm{\scriptsize~\cite{Gillick:2009:NAACL}} & 37.31 & 9.36 & 13.12 \\
\hline
DPP{\scriptsize~\cite{Kulesza:2012}}$\dagger$ & 38.10 & 9.14 & 13.40 \\
DPP-Caps{\scriptsize~\cite{Cho:2019}}  & 38.25 & 9.22 & 13.40 \\
DPP-Caps-Comb{\scriptsize~\cite{Cho:2019}}  & 39.35 & 10.14 & 14.15 \\
DPP-BERT (ours) & {38.14} & {9.30} & {13.47} \\
DPP-BERT-Comb 64 (ours) & {38.78} & {9.78} & {14.04} \\
DPP-BERT-Comb 128 (ours) & \textbf{39.05} & \textbf{10.23} & \textbf{14.35} \\
\hline
\end{tabular}
\end{small}
\caption{Results on the DUC-04 dataset evaluated by ROUGE. 
$\dagger$ indicates our reimplementation of Kulesza and Taskar~\shortcite{Kulesza:2012} system.
}
\label{tab:results_duc04}
\vspace{-0.1in}
\end{table}

\section{Experiments}

In this section we describe the dataset used to train the BERT-\emph{sim} and BERT-\emph{imp} models, benchmark datasets for multi-document summarization, and experimental settings. 
Our system shows competitive results comparing to state-of-the-art methods. 
Example summaries are provided to demonstrate the effectiveness of the proposed method.

\subsection{Dataset}
\label{sec:data}

\noindent\textbf{CNN / DailyMail}\quad
This dataset~\cite{Hermann:2015} is utilized to train the BERT-\emph{sim} and BERT-\emph{imp} models.
For BERT-\emph{sim}, we pair each human summary sentence with its most similar document sentence to create a positive instance; negative instances are randomly sampled sentence pairs. 
For BERT-\emph{imp}, the most similar document sentence receives a label of 1; randomly sampled sentences are labelled as 0. In total, our training / dev / test sets contain 2,084,798 / 105,936 / 86,144 sentence pairs and the instances are balanced.

\vspace{0.05in}
\noindent\textbf{DUC/TAC}\quad
We evaluate our DPP approach (\S\ref{sec:dpp}) on multi-document summarization datasets including DUC and TAC~\cite{Over:2004,Dang:2008}.
The task is to generate a summary of 100 words from a collection of news articles.
We report ROUGE F-scores~\cite{Lin:2004}\footnote{with options \textsf{-n 2 -m -w 1.2 -c 95 -r 1000 -l 100}} on DUC-04 (trained on DUC-03) and TAC-11 (trained on TAC-08/09/10) following standard settings~\cite{Hong:2014}.
Ground-truth extractive summaries used in DPP training are obtained from Cho et al.~\shortcite{Cho:2019}.

\subsection{Experiment Settings}
We implement our system using TensorFlow on an NVIDIA 1080Ti GPU. 
We consider the maximum length of a sentence to be 64 or 128 words. 
The batch size is 64 for the 64 max sentence length and 32 for 128. 
We use Adam optimizer~\cite{Kingma:2015} with the default setting and set learning rate to be 2e-5. 
We train BERT-\emph{imp} and BERT-\emph{sim} on CNN/DM.
The prediction accuracy of BERT-\emph{sim} and BERT-\emph{imp} (with length-128) are respectively 96.11\% and 69.05\%. 
Similar results are observed with length-64: 95.79\% and 69.63\%.

\subsection{Summarization Results}

We compare our system with strong summarization baselines (Table~\ref{tab:results_duc04} and~\ref{tab:results_tac11}).
\textit{SumBasic}~\cite{Vanderwende:2007}, \textit{KL-Sum}~\cite{Haghighi:2009}, and \textit{LexRank}~\cite{Erkan:2004} are extractive approaches;  \textit{Opinosis}~\cite{Ganesan:2010}, \textit{Extract+Rewrite}~\cite{Song:2018}, and \textit{Pointer-Gen}~\cite{See:2017} are abstractive methods; \textit{ICSISumm}~\cite{Gillick:2009:UTD} is an ILP-based summarization method; and
\textit{DPP-Caps-Comb}, \textit{DPP-Caps} are results combining DPP and capsule networks reported by Cho et al.~\shortcite{Cho:2019} w/ and w/o using sentence TF-IDF similarity ($C_{i,j}$).

We experiment with variants of our DPP model: \textit{DPP-BERT}, \textit{DPP-BERT-Combined}. 
The former utilizes the outputs from BERT-\emph{sim} and BERT-\emph{imp} to compute $S_{ij}$ and $q_{i}$, whereas the latter combines BERT-\emph{sim} output with sentence TF-IDF similarity  ($C_{i,j}$), and concatenates BERT-\emph{imp} features with linguistically informed features.

Our DPP methods outperform both extractive and abstractive baselines, indicating the effectiveness of optimization-based methods for extractive multi-document summarization.
Furthermore, we observe that \textit{DPP-BERT-Combined} yields the best performance, achieving 10.23\% and 11.06\% F-scores respectively on DUC-04 and TAC-11.
This finding suggests that sentence similarity scores and importance features from the \textit{DPP-BERT} system and TF-IDF based features can complement each other to boost system performance. 
We conjecture that TF-IDF sentence vectors are effective at representing topical terms (e.g., \emph{3 million}), thus helping DPP better select representative sentences. Another observation is that \textit{DPP-BERT} and \textit{DPP-BERT-Combined} consistently outperform \textit{DPP-Caps} and \textit{DPP-Caps-Comb}, indicating its excellence for DPP-based summarization.

In Table~\ref{tab:example_summaries} we show example system summaries and a human-written reference summary.
\textit{DPP-BERT} and \textit{DPP-BERT-Combined} both are capable of selecting a balanced set of representative and diverse summary sentence from multi-documents.
\textit{DPP-BERT-Combined} selects more relevant sentences than \textit{DPP-BERT} comparing to the human summary, leading to better ROUGE scores.

\begin{table}[!t]
\setlength{\tabcolsep}{4.9pt}
\renewcommand{\arraystretch}{1.2}
\centering
\begin{small}
\begin{tabular}{|l|rrr|}
\hline
& \multicolumn{3}{c|}{\textbf{TAC-11}}\\
\textbf{System} & \textbf{R-1} & \textbf{R-2} & \textbf{R-SU4} \\
\hline
\hline
Opinosis{\scriptsize~\cite{Ganesan:2010}} & 25.15 & 5.12 & 8.12\\
Extract+Rewrite{\scriptsize~\cite{Song:2018}} & 29.07 & 6.11 & 9.20\\
Pointer-Gen{\scriptsize~\cite{See:2017}} & 31.44 & 6.40 & 10.20\\
SumBasic{\scriptsize~\cite{Vanderwende:2007}} & 31.58 & 6.06 & 10.06\\
KLSumm{\scriptsize~(Haghighi et al., 2009)\nocite{Haghighi:2009}} & 31.23 & 7.07 & 10.56 \\
LexRank{\scriptsize~\cite{Erkan:2004}} & 33.10 & 7.50 & 11.13 \\
\hline
DPP{\scriptsize~\cite{Kulesza:2012}}$\dagger$ & 36.95 & 9.83 & 13.57 \\
DPP-Caps{\scriptsize~\cite{Cho:2019}} & {36.61} & {9.30} & {13.09} \\
DPP-Caps-Comb{\scriptsize~\cite{Cho:2019}} & {37.30} & {10.13} & {13.78} \\
DPP-BERT (ours) & {37.04} & {10.18} & {13.79} \\
DPP-BERT-Comb 64 (ours) & {38.46} & {10.79} & {14.45} \\
DPP-BERT-Comb 128 (ours) & \textbf{38.59} & \textbf{11.06} & \textbf{14.65} \\
\hline
\end{tabular}
\end{small}
\caption{ROUGE results on the TAC-11 dataset. 
}
\label{tab:results_tac11}
\vspace{-0.1in}
\end{table}

\begin{table}[!t]
\setlength{\tabcolsep}{5pt}
\renewcommand{\arraystretch}{1.1}
\begin{scriptsize}
\begin{fontppl}

\begin{minipage}[b]{0.5\hsize}\centering
\begin{tabular}{|p{2.85in}|}
\toprule
\textbf{Human Reference Summary}\\[2mm]
\textbullet\, On March 1, 2007, the Food/Drug Administration (FDA) started a broad safety review of children's cough/cold remedies. \\[1.mm]

\textbullet\, They are particularly concerned about use of these drugs by infants. \\[1.mm]

\textbullet\, By September 28th, the 356-page FDA review urged an outright ban on all such medicines for children under six. \\[1.mm]

\textbullet\, Dr. Charles Ganley, a top FDA official said ``We have no data on these agents of what's a safe and effective dose in Children.'' The review also stated that between 1969 and 2006, 123 children died from taking decongestants and antihistimines. \\[1.mm]

\textbullet\, On October 11th, all such infant products were pulled from the markets. \\[3mm]
\bottomrule
\toprule
\textbf{DPP-BERT Summary}\\[2mm]
\textbullet\, The petition is far from the first warning about children using the medicines.\\[1.mm]

\textbullet\, The FDA will formally consider revising labeling at a meeting scheduled for Oct. 18-19.\\[1.mm]

\textbullet\, Federal drug regulators have started a broad review of the safety of popular cough and cold remedies meant for children, a top official said Thursday.\\[1.mm]

\textbullet\, Similarly, hydrocodone has never been shown to be safe and effective in children, and its dangers as a powerful and potentially addictive narcotic are clear.\\[1.mm]
\bottomrule
\toprule
\textbf{DPP-BERT-Combined Summary}\\[1mm]
\textbullet\, The U.S. government is warning parents not to give cough and cold medicines to children under 2 without a doctor's order, part of an overall review of the products' safety and effectiveness for youngsters.\\[1.mm]

\textbullet\, Drug makers on Thursday voluntarily pulled kids' cold medicines off the market less than two weeks after the U.S. government warned of potential health risks to infants.\\[1.mm]

\textbullet\, Safety experts for the Food and Drug Administration urged the agency on Friday to consider an outright ban on over-the-counter, multi-symptom cough and cold medicines for children under 6.\\[1.mm]

\textbullet\, In high doses, cold medicines can affect the heart's electrical system, leading to arrhythmias.\\[1.mm]
\bottomrule
\end{tabular}
\end{minipage}

\end{fontppl}
\end{scriptsize}
\caption{Example system summaries and their human reference summary.
Sentences selected by DPP-BERT-Combined are more similar to the human summary than those of DPP-BERT; 
both include diverse sentences. 
}
\label{tab:example_summaries}
\end{table}

\section{Conclusion}
In this paper we describe a novel approach using determinantal point processes for extractive multi-document summarization. 
Our DPP+BERT models harness the power of deep contextualized representations and optimization to achieve outstanding performance on multi-document summarization benchmarks.
Our analysis further reveals that, despite the success of deep contextualized representations, it remains necessary to combine them with surface indicators for effective identification of summary-worthy sentences.

\section*{Acknowledgments}

We are grateful to the anonymous reviewers for their helpful suggestions.
This research was supported in part by the National Science Foundation grant IIS-1909603.

\bibliography{summ,fei,abs_summ,emnlp-ijcnlp-2019}
\bibliographystyle{acl_natbib}

\end{document}